\documentclass[runningheads]{llncs}
\usepackage[T1]{fontenc}
\usepackage{graphicx}
\usepackage[table,xcdraw]{xcolor}
\usepackage{subcaption}
\usepackage{comment}
\usepackage{makecell}
\usepackage{caption}
\usepackage{booktabs} 
\usepackage{cite} 
\usepackage{hyperref}
\begin{document}
\title{Artificial Conversations, Real Results: Fostering Language Detection with Synthetic Data}
%
%
\author{Fatemeh Mohammadi\inst{1}\orcidID{\href{https://orcid.org/0000-0002-3100-7569}{0000-0002-3100-7569}} \and
Tommaso Romano\inst{1}\and
Samira Maghool\inst{1}\orcidID{\href{https://orcid.org/0000-0001-8310-2050}{0000-0001-8310-2050}} \and
Paolo Ceravolo\inst{1}\orcidID{\href{https://orcid.org/0000-0002-4519-0173}{0000-0002-4519-0173}}}
\authorrunning{F. Mohammadi et al.}
%
\institute{University of Milan, Milan, Italy}
\titlerunning{Artificial Conversations, Real Results}
\maketitle              
\begin{abstract}
Collecting high-quality training data is essential for fine-tuning Large Language Models (LLMs). However, acquiring such data is often costly and time-consuming, especially for non-English languages such as Italian. Recently, researchers have begun to explore the use of LLMs to generate synthetic datasets as a viable alternative. This study proposes a pipeline for generating synthetic data and a comprehensive approach for investigating the factors that influence the validity of synthetic data generated by LLMs by examining how model performance is affected by metrics such as prompt strategy, text length and target position in a specific task, i.e. inclusive language detection in Italian job advertisements. Our results show that, in most cases and across different metrics, the fine-tuned models trained on synthetic data consistently outperformed other models on both real and synthetic test datasets. The study discusses the practical implications and limitations of using synthetic data for language detection tasks with LLMs.

\end{abstract}
\section{Introduction}

In recent years, Large Language Models (LLMs) have received considerable attention in language recognition tasks. However, the effectiveness of these models largely depends on appropriate fine-tuning procedures, which require access to large, diverse and high-quality datasets for training and evaluation. Obtaining such datasets poses significant challenges, including data scarcity, privacy, unbalanced datasets, lack of edge cases, and the high costs associated with data collection and annotation \cite{liu2024best}.

Synthetic data has been proposed as a potential solution to address certain challenges associated with language detection tasks, including limited data availability \cite{founta2018large} and the psychological impact on annotators \cite{riedl2020downsides}. Synthetic data helps to overcome these limitations by offering greater control over data properties, allowing tailored augmentation for specific tasks, such as testing models under different conditions or rare scenarios~\cite{Prakash2019structured}. 

It also accelerates the iterative process of model development by providing readily available and customisable datasets~\cite{Le2020machine}. In addition, synthetic data helps mitigate biases present in real-world datasets~\cite{maghool2023enhancing} and improves model generalisation by exposing algorithms to a wider range of input variations~\cite{miao2024domaindiff}. This approach is particularly valuable in areas such as healthcare~\cite{Bellandi2022methodology}, complex systems~\cite{maghool2020epidemic}, and natural language processing~\cite{Eigenschink2023deep}, where obtaining high-quality labelled data is often difficult or resource-intensive.

The capabilities of language models for most language detection tasks have been extensively discussed \cite{sen2024hatetinyllm, guo2023investigation, di2024explanation, he2024guardians}. However, an investigation of the utility of LLMs for detecting inclusive language is still lacking.  In particular, an in-depth evaluation of the capabilities of such models, e.g. their ability to achieve high performance even when fine-tuned using synthetic data. Therefore, in this paper we aim to address the challenges associated with acquiring high quality training data for fine-tuning LLMs, especially in resource-constrained settings such as non-English languages. In this paper, we address these challenges through the following contributions:
\begin{enumerate}
\item Proposing a synthetic data generation pipeline to address data scarcity in resource-constrained settings.  
\item Outline a workflow that involves fine-tuning an LLM on synthetic training data, followed by inference with fine-tuned and pre-trained models on synthetic test data to evaluate the effectiveness of synthetic data.  
\item Focusing on inclusive language detection, an under-researched and challenging task, especially in gendered languages such as Italian.  
\item Demonstrate the potential of synthetic data as a cost-effective, scalable solution by showing that fine-tuned models trained on synthetic data outperform other models on both real and synthetic test data.  
\end{enumerate}

We review the related work and background in Sect. \ref{sec:background}. We then present our proposed methodology in Sect. \ref{sec:methodology}, followed by the evaluation and discussion of the results in Sect. \ref{res}. Finally, we conclude the paper in Sect. \ref{sec:conclusion} with some remarks and plans for the future.

\section{Background and Related Works}\label{sec:background}

\subsection{Synthetic data generation}
Synthetic data generation has become an essential approach to mitigate challenges related to data scarcity, privacy, and the need for diverse datasets when training machine learning models \cite{lu2023machine}. Different techniques, including Generative Adversarial Networks (GANs), Variational Autoencoders (VAEs), and LLMs, offer different capabilities tailored to specific data types. GANs are particularly effective for generating tabular data, while VAEs are widely used for generating synthetic images and tabular datasets \cite{goyal2024systematic}. 

For synthetic text data, LLMs are the most suitable choice. These models produce coherent, contextually accurate text that is virtually indistinguishable from human-written content, making them ideal for natural language processing tasks \cite{li2023synthetic}. By providing well-crafted prompts or instructions, LLMs can generate diverse and realistic textual data, such as dialogues, narratives, and domain-specific content \cite{goyal2024systematic}. Such synthetic data can be used for a variety of purposes, including fine-tuning LLMs, increasing the diversity of datasets, and simulating scenarios for testing and development.

\subsection{Using synthetic data for language detection tasks}
Several studies have used synthetic data for language detection tasks. Authors in \cite{casula2024don} explored the possibility of replacing existing datasets in English for abusive language detection with synthetic data obtained by rewriting original texts with an instruction-based generative model. They showed that such data can be effectively used to train a classifier whose performance is equal to, and sometimes better than, a classifier trained on original data. In another study, the authors show that for intent detection tasks, i.e. a classification task that involves determining the underlying goal behind a user query, synthetic data can effectively predict the performance of different approaches~\cite{maheshwari2024efficacy}. In addition, \cite{khullar2024hate} demonstrated the generation of synthetic data for hate speech detection in low-resource languages, such as Hindi and Vietnamese, using various methods. These include automatic Machine Translation (MT) of hateful posts from a high-resource language and Contextual Entity Substitution (CES). The CES method uses a small set of examples in a high-resource language, such as English, and heuristically replaces the target person or group in the high-resource context with potential hateful targets or groups relevant to the target language context. Their results show that a model trained on synthetic data performs comparably to, and in some cases outperforms, a model trained only on the examples available in the target domain.

Despite existing research on the use of synthetic data for tasks such as detecting hate speech and abusive language, there is a notable lack of focus on inclusive language. Specifically, in the context of generating synthetic data for job advertisements, to our knowledge only one study has been developed \cite{magron2024jobskape}. This study presented \textit{SkillSkape}, an open-source synthetic dataset of job advertisements designed for skill matching tasks rather than comprehensive language detection. Given the gendered nature of the Italian language and the paucity of research addressing this issue, our study offers significant novelty and fills an important gap in the field.

\subsection{Fine-tuning LLMs using synthetic data}
The use of LLMs to generate synthetic data is growing; however, there is limited research on fine-tuning LLMs specifically using synthetic data. One notable study \cite{li2024synthetic} introduced Generalised Instruction Tuning (GLAN), a method for generating synthetic data tailored to instruction tuning tasks. Extensive experiments on the Mistral model showed that training with GLAN enabled the model to excel in several domains, including mathematical reasoning, coding, academic exams, logical reasoning, and general instruction following. Remarkably, this was achieved without the use of task-specific training data for these applications.
The study most closely related to our methodology is \cite{zhezherau2024hybrid}, which fine-tuned a model using a combination of real and synthetic data. However, their focus is different, as their synthetic data relate to therapy sessions. Their experimental results showed that the hybrid model consistently outperformed others in specific vertical applications, achieving superior performance across all metrics. Additional tests confirmed the hybrid model's enhanced adaptability and contextual understanding in different scenarios. These results highlight the potential of combining real and synthetic data to improve the robustness and contextual sensitivity of LLMs, particularly for domain-specific and specialised applications.

\section{Methodology} \label{sec:methodology}
\begin{figure}
\centering
\includegraphics[width=0.8\textwidth]{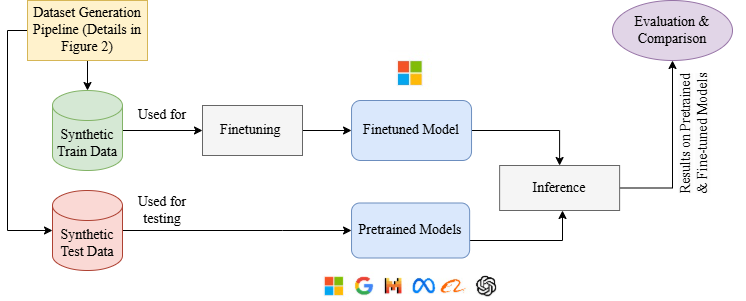}
\caption{An overview of the whole framework} \label{fig1}
\end{figure}

This section outlines our proposed approach for developing a framework driven by LLMs, designed to generate synthetic data and evaluate it using different prompt strategies across different language models. In this study, we used this framework to detect non-inclusive language in Italian job advertisements. Our approach consists of five main parts: (i) creation of a synthetic dataset by combining real and generated data, (ii) study of different prompting techniques and their impact on the performance of pre-trained models, (iii) fine-tuning of a model on the synthetic data, (iv) inference using our fine-tuned model and other pre-trained models, on both synthetic and real seed test datasets, and finally (v) a comprehensive comparative analysis of the results, including performance under different parametrization. Figure \ref{fig1} shows an overview of the whole framework. 

\subsection{Synthetic data generation}

Synthetic data generation is a key part of our methodology, enabling the creation of large datasets for both fine-tuning and evaluation purposes. A major challenge in this process is the risk of repetition within the dataset, where variation is limited to minor changes in the words that replace the placeholders. To mitigate this, we implemented a novel strategy: pre-splitting the templates into separate training and test sets prior to data generation. By separating the templates at the outset, we ensure that the test set remains sufficiently distinct from the training set, reducing the overlap of similar sentence structures. This approach minimises overfitting and improves the overall robustness and performance of our fine-tuned model.

\begin{figure}
\includegraphics[width=\textwidth]{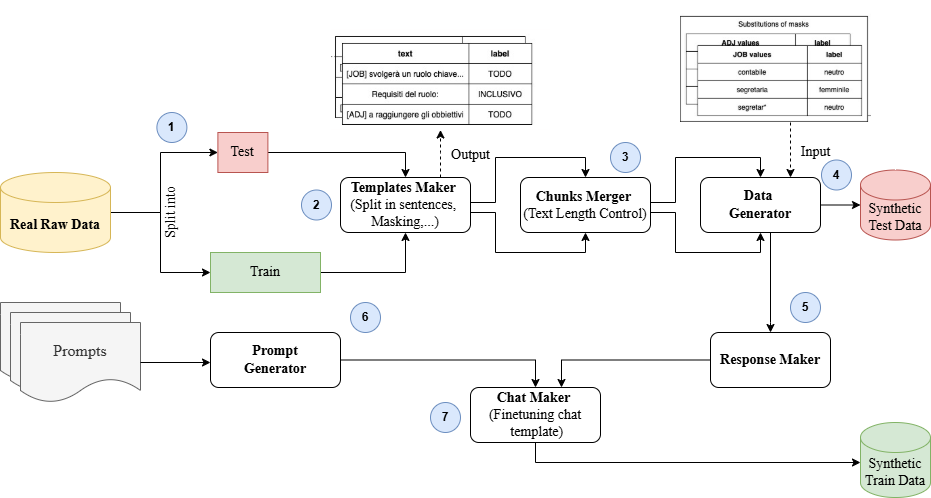}
\caption{Synthetic data generation workflow} \label{fig2}
\end{figure}

As shown in Figure \ref{fig2}, we adopted the conventional 70-30 split for training and testing the models, allocating 70\% of the dataset for training and the remaining 30\% for evaluation. This widely used approach provides the model with sufficient data for effective learning, while ensuring that an independent test set is available for unbiased performance evaluation. By using this partitioning strategy, we achieve a balance between optimising the model's learning process and validating its generalisation capabilities, thereby reducing the risk of data leakage. This ensures a robust and reliable evaluation of the model's performance on previously unseen data.

The process starts with real data, which is deconstructed into individual sentences (No. 1).  Sentences containing words that can be masked are identified and reused as templates for dataset construction. Each sentence is given a binary label: {\tt TODO} for sentences with maskable words that require further processing, and {\tt INCLUSIVE} for sentences that are inherently neutral and cannot be discriminated. For example, a sentence like ``Sarai {\tt [VERB]} per un colloquio conoscitivo'' is labelled {\tt TODO}, while a neutral sentence like ``Descrizione del ruolo:'' is categorised as {\tt INCLUSIVE} (No. 2). This labelling system allows a clear distinction between maskable and non-maskable sentences, thus ensuring an organised and targeted approach to the creation of the dataset (\textit{Template Maker} module). 

Research has shown that text length plays a crucial role in shaping the performance and behaviour of LLMs, affecting aspects such as coherence, contextual understanding and generation quality \cite{baillargeon2024assessing}. A notable advantage of our synthetic data generation pipeline is the inclusion of a \textit{chunk merger} module (No. 3), which allows precise control over text length. This feature allows the creation of synthetic data sets of different lengths, facilitating in-depth analysis of how text length affects LLM performance. We create templates containing placeholders for \textit{job titles}, \textit{work-related adjectives} and \textit{verbs}, categorised \textit{by gender}. In the next step (No. 4), we replace the placeholders in the annotated dataset with a corresponding word from a substitution dataset, where each vocabulary is labelled as {\tt neutral}, {\tt masculine} or {\tt feminine}. This substitution process generates a large number of possible text combinations (\textit{Chunk Merger} module).  

After generating synthetic test data, the next step is to generate labelled training data, which will be used for fine-tuning the LLM. To achieve this, the \textit{Response Maker} module (No. 5) is used to assign labels to each sentence generated by the \textit{Data Generator} module. Sentences containing only one label {\tt masculine} or {\tt feminine} are classified as {\tt NONINCLUSIVE}, while those containing only {\tt neutral} elements are classified as {\tt INCLUSIVE}. For example, if the placeholder {\tt [JOB]} in the template sentence ``{\tt[JOB]} svolgerà un ruolo chiave...'' is replaced by {\tt insegnante} (teacher), the sentence is marked as {\tt INCLUSIVE}, because {\tt insegnante} is a gender-neutral term in Italian. Conversely, replacing {\tt [JOB]} with {\tt infermiere} (nurse) results in a label of {\tt NONINCLUSIVE}, as {\tt infermiere} is a masculine term that excludes female candidates. This systematic labelling ensures accurate identification of inclusive language within the dataset.

To fine-tune the LLM using chat template data, we use the \textit{Prompt Generator} module along with the labelled sentences from the previous step. By feeding these two inputs into the \textit{Chat Maker} module (No. 7), we generate 10,424 rows of data, which will serve as the training data set for fine-tuning the LLM. The details of the \textit{Prompt Generator} module will be discussed in the next section.

\subsection{Prompts generation based on different approaches}\label{PG}
One of the main aims of this study is to investigate how different prompting methods affect the responses produced by LLMs, with the overall aim of optimising response quality. Recent research has shown that even small variations in prompting - such as rephrasing or changing the structure - can have a significant impact on LLM performance~\cite{beurer2023prompting}. For example, strategies such as encouraging step-by-step reasoning or rephrasing objectives can lead to markedly different outcomes~\cite{reynolds2021prompt}. To streamline this exploration, we are implementing an automated system for designing and managing prompts using a modular prompting framework. This framework includes methods such as \textit{zero-shot learning} (ZSL), \textit{few-shot learning} (FSL) and ZSL with \textit{chain-of-thought} strategy, which we call (ZSLCOT) prompting~\cite{lee2024applying}. Using these prompting methods, we were able to generate four different prompts: two for ZSL and one for each of the other strategies. The results provide valuable insights into how to effectively design prompts to maximise LLM performance and response quality.
\subsection{LLM fine-tuning}
We fine-tuned a pre-trained language model using the \textit{Unsloth} library, known for its powerful fine-tuning capabilities and accelerated processing speed \cite{finetuneunsloth}. For this study, we chose the \textit{Phi3-mini} model, a compact and cost-effective architecture with 29,884,416 trainable parameters. To further optimise efficiency, we used a 4-bit quantized version of the pre-trained model, which significantly reduced computational requirements and processing time without compromising performance.
The fine-tuning dataset consisted of 5,712 synthesised samples formatted as chat data containing questions, text and responses. These samples were tokenised using a custom chat template designed specifically for the \textit{Phi-3} architecture to ensure compatibility and maximise the effectiveness of the fine-tuning process. 

The tuning used \textit{Parameter-Efficient Fine-Tuning} (PEFT) with a LoFTQ configuration~\cite{li2023loftq}, which allows tuning without changing all model parameters, making the process more resource efficient. Using \textit{SFTTrainer} from the Hugging Face library~\cite{jain2022hugging}, a single epoch of 360 training steps was run on a Tesla T4 GPU with 14.748 GB of RAM, achieving a speed of 0.28 iterations per second and completing in 26.55 minutes. The resulting fine-tuned model was then uploaded to a model hub and made available for further use. 


\subsection{Inference using fine-tuned and pre-trained models}
The inference process defined in this study refers to the procedure by which both pre-trained and fine-tuned language models generate responses based on input data. This process is applied consistently to both the synthetic test dataset and the manually annotated real-world dataset, ensuring a consistent evaluation framework. This dual evaluation approach minimises the risk of overfitting by validating model performance on different data sources. Answers are generated using the automatically generated prompts, as described in Section \ref{PG}.

We compared the results obtained by our fine-tuned model with five different LLMs: \textit{LLaMA} 3 7B from Meta, \textit{Phi-3-mini} 3.8B from Microsoft, \textit{Mistral} 7B, \textit{Qwen} 2 7B from Alibaba and \textit{Gemma} 2 9B from Google. The \textit{Phi-3-mini} used in the comparison is different from the \textit{Phi-3-mini} we have fine-tuned. 
Our data collection yielded a substantial dataset of $10,424$ responses, for a total of $62,544$ data points. This large dataset enables a thorough evaluation of each model, prompt, and inferred label for the examples.
\subsection{Comprehensive comparative analysis}
In this paper, we present a comprehensive evaluation of different LLMs using a range of metrics and tasks, focusing specifically on the detection of non-inclusive language in job advertisements. The detection of non-inclusive language is framed as a binary classification problem, and we evaluate the models and tasks from several perspectives. Our evaluation includes the following key aspects. (i) The \textit{structure of the top responses} generated by different LLMs, highlighting how well the output matches the provided prompts. (ii) A comparison of the \textit{accuracy of different prompt strategies} applied to both synthetic and real seed datasets, identifying the most effective models and prompts for each. (iii) Having identified the best performing prompt for each model, we evaluate \textit{model performance} using these optimised prompts. We use \textit{precision}, \textit{recall}, \textit{specificity}, \textit{F1 score} and \textit{accuracy} as the primary metrics, given their widespread use in classification tasks. However, due to the issue of data imbalance, often cited in the literature as a source of metric skewness and potential bias, we also include \textit{balanced accuracy} (bACC) as a more reliable metric. Balanced accuracy is particularly valuable for unbalanced datasets, as it adjusts for differences in class distributions, providing a more accurate reflection of model performance across both classes \cite{brodersen2010balanced}. 
Further analysis includes (iv) investigating the effect of \textit{text length} and target word position on model performance, both in synthetic and real data, to gain deeper insights into how these factors influence accuracy. 

\section{Results and discussion}\label{res}
\subsection{Evaluation the structure of the produced responses}
After generating responses using LLMs, the first step is to pre-process the output to ensure that the responses are standardised. The primary objective is to extract the binary labels - {\tt INCLUSIVE} and {\tt NON-INCLUSIVE} - from these responses. This pre-processing step is essential because the generated output often contains extraneous information, such as reasoning and explanations, which must be removed to produce a clean set of responses for accurate analysis.

\begin{figure}
\includegraphics[width=\textwidth]{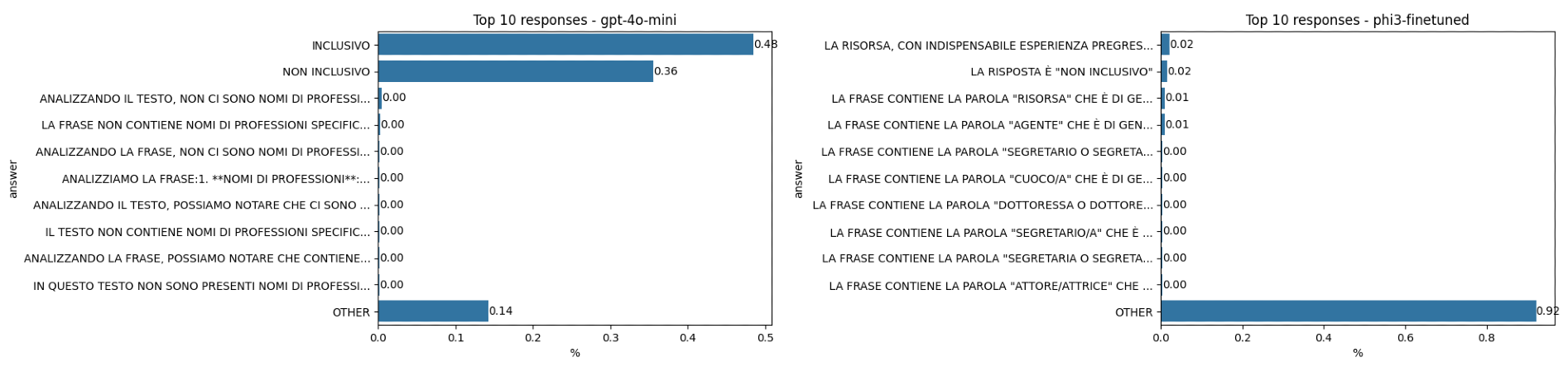}
\caption{Top 10 responses produced by LLMs} \label{fig5}
\end{figure}

Figure \ref{fig5} shows the top 10 responses from \textit{gpt-4o-mini} and fine-tuned \textit{Phi3} models as examples evaluated on the same test dataset. While the pre-trained models such as \textit{gpt-4o-mini} produce output in the correct format, the fine-tuned model often introduces additional noise such as explanations or special characters.  
To handle the varying outputs, several preprocessing functions were implemented to extract the desired labels. As a result, 99.90\% of the responses were successfully transformed into the desired format through automated preprocessing, with less than 0.1\% of responses falling outside the expected format.

Another key observation is the class imbalance in the responses generated. The distribution of responses generated by the LLMs for the synthetic dataset shows that {\tt INCLUSIVE} labels appear twice as often as {\tt NON INCLUSIVE} labels. This imbalance supports the choice of \textit{balanced accuracy} (bACC) as a more appropriate evaluation metric.
\subsection{Evaluation of the prompting strategies}
As discussed in the methodology section, the quality of the prompt is crucial for obtaining optimal responses from both pre-trained and fine-tuned LLMs. To save time and computational resources, we decided to use a single prompt strategy during the inference process, so before starting, we compare the different prompts automatically generated by our pipeline based on their accuracy to select the best one. Table \ref{tab:combined_tables} shows the accuracy of our four tested prompts for the synthetic \ref{tab:subtable_a} and seed \ref{tab:subtable_b} datasets, respectively. 
\begin{table}[ht]
    \centering

    \begin{subtable}[t]{0.8\textwidth} 
        \centering
        \resizebox{!}{1.2cm}
{
        \begin{tabular}{l@{\hspace{0.5cm}}|c@{\hspace{0.5cm}}c@{\hspace{0.5cm}}c@{\hspace{0.5cm}}c}
        \hline
        \multicolumn{1}{c|}{{Model}} &  FSL\#0 & ZSL\#0 & ZSL\#1 & ZSLCOT\#0\\
        \hline
        phi3\_finetuned &
        \cellcolor[HTML]{E2EFDA}\textbf{0.973} &
        \cellcolor[HTML]{E2EFDA}\textbf{0.976} &
        \cellcolor[HTML]{E2EFDA}\textbf{0.921} &
        \cellcolor[HTML]{E2EFDA}\textbf{0.991}\\
        gpt\_40\_mini & \textbf{0.888} & 0.810          & 0.631 & 0.820          \\ 
        phi3          & \textbf{0.508} & 0.504          & 0.475 & 0.499          \\ 
        llama3        & \textbf{0.563} & 0.502          & 0.546 & 0.526          \\ 
        mistral       & 0.503          & 0.510          & 0.502 & \textbf{0.520} \\ 
        gemma2        & \textbf{0.598} & 0.553          & 0.431 & 0.537          \\ 
        qwen2         & 0.511          & \textbf{0.580} & 0.471 & 0.579          \\ 
        \hline
        \end{tabular}}
        \caption{Synthetic dataset}
        \label{tab:subtable_a}
    \end{subtable}
    \begin{subtable}[t]{0.8\textwidth} 
        \centering
        \resizebox{!}{1.2cm}
{
        \begin{tabular}{l@{\hspace{0.5cm}}|c@{\hspace{0.5cm}}c@{\hspace{0.5cm}}c@{\hspace{0.5cm}}c}
        \hline
        \multicolumn{1}{c|}{{Model}}  &  FSL\#0 & ZSL\#0 & ZSL\#1 & ZSLCOT\#0\\
        \hline
        phi3\_finetuned &
        \cellcolor[HTML]{E2EFDA}\textbf{0.642} &
        \cellcolor[HTML]{E2EFDA}\textbf{0.702} &
        \cellcolor[HTML]{E2EFDA}\textbf{0.647} &
        \cellcolor[HTML]{E2EFDA}\textbf{0.677}\\
        gpt\_40\_mini & 0.565 & \textbf{0.647}          & 0.595 & 0.585          \\ 
        phi3          & \textbf{0.512} & 0.500          & 0.502 & 0.500          \\ 
        llama3        & 0.512          & \textbf{0.525} & 0.519 & 0.515          \\ 
        mistral       & 0.500          & \textbf{0.542} & 0.490 & 0.501          \\ 
        gemma2        & \textbf{0.545} & 0.444          & 0.516 & 0.535          \\ 
        qwen2         & 0.523          & 0.525          & \textbf{0.542} & 0.537 \\ 
        \hline
        \end{tabular}}
        \caption{Seed dataset}
        \label{tab:subtable_b}
    \end{subtable}

    \caption{The comparison of accuracy between prompts for different models on synthetic and seed datasets.}
    \label{tab:combined_tables}
\end{table}

The results show that the FSL strategy generally performs better on synthetic data, achieving the highest accuracy in five of the seven models tested. In particular, the fine-tuned model shows the best performance of all strategies on the synthetic test data. On the real seed dataset, our fine-tuned model also shows superior accuracy compared to the others. Overall, both the FSL and ZSL methods are effective on this dataset, with FSL outperforming three models and ZSL outperforming four of the seven models tested.
\subsection{Evaluation of the models' performance}
Having identified the best performing prompt for each model, we performed a comparative analysis of the key performance metrics. The results presented in Table \ref{tab:model_performance} show that our fine-tuned model outperforms on all metrics for the synthetic test data (\ref{tab:subtable1_a}) and on most metrics (four out of six) for the real seed data (\ref{tab:subtable2_b}). This shows that training LLMs with synthetic data can be highly effective on real data, highlighting the potential of synthetic data for language detection tasks, even in complex contexts such as non-inclusive language. 
\begin{table}[ht]
    \centering

    \begin{subtable}[t]{\textwidth} 
        \centering
        \resizebox{!}{1.1cm}
{
        \begin{tabular}{l@{\hspace{0.3cm}}|c@{\hspace{0.3cm}}c@{\hspace{0.3cm}}c@{\hspace{0.3cm}}c@{\hspace{0.3cm}}c@{\hspace{0.3cm}}c}
        \hline
        \multicolumn{1}{c|}{{Model}} &
        Recall &
        Specificity &
        Accuracy &
        bACC &
        Precision &
        F1-score \\ \hline
        phi3\_finetuned\_zslcot\#0 &
          \cellcolor[HTML]{E2EFDA}\textbf{0.990} &
          \cellcolor[HTML]{E2EFDA}\textbf{0.992} &
          \cellcolor[HTML]{E2EFDA}\textbf{0.991} &
          \cellcolor[HTML]{E2EFDA}\textbf{0.991} &
          \cellcolor[HTML]{E2EFDA}\textbf{0.997} &
          \cellcolor[HTML]{E2EFDA}\textbf{0.993} \\
        gpt\_40\_mini\_fsl\#0 & 0.797          & 0.979 & 0.851 & 0.888 & \textbf{0.989} & 0.883 \\
        phi3\_fsl\#0          & \textbf{0.972} & 0.045 & 0.696 & 0.508 & 0.707          & 0.818 \\
        llama3\_fsl\#0        & \textbf{0.950} & 0.177 & 0.720 & 0.563 & 0.732          & 0.827 \\
        mistral\_zslcot\#0    & \textbf{0.777} & 0.262 & 0.626 & 0.520 & 0.718          & 0.746 \\
        gemma2\_fsl\#0        & 0.526          & 0.671 & 0.569 & 0.598 & \textbf{0.791} & 0.632 \\
        qwen2\_zsl\#1         & \textbf{0.930} & 0.230 & 0.722 & 0.580 & 0.741          & 0.824 \\ \hline
        \end{tabular}}
        \caption{Synthetic dataset}
        \label{tab:subtable1_a}
    \end{subtable}


    \begin{subtable}[t]{\textwidth} 
        \centering
        \resizebox{!}{1.1cm}
{
        \begin{tabular}{l@{\hspace{0.3cm}}|c@{\hspace{0.3cm}}c@{\hspace{0.3cm}}c@{\hspace{0.3cm}}c@{\hspace{0.3cm}}c@{\hspace{0.3cm}}c}
        \hline
        \multicolumn{1}{c|}{{Model}} & Recall         & Specificity    & Accuracy & bACC  & Precision      & F1-score \\ \hline
        phi3\_finetuned\_zslcot\#0 &
          \cellcolor[HTML]{E2EFDA}\textbf{0.824} &
          \cellcolor[HTML]{E2EFDA}0.581 &
          \cellcolor[HTML]{E2EFDA}\textbf{0.713} &
          \cellcolor[HTML]{E2EFDA}\textbf{0.702} &
          \cellcolor[HTML]{E2EFDA}0.700 &
          \cellcolor[HTML]{E2EFDA}\textbf{0.757} \\
        gpt\_40\_mini\_fsl\#0               & 0.549          & \textbf{0.744} & 0.638    & 0.647 & \textbf{0.718} & 0.622    \\
        phi3\_fsl\#0                        & \textbf{1.000} & 0.023          & 0.553    & 0.512 & 0.548          & 0.708    \\
        llama3\_fsl\#0                      & \textbf{0.980} & 0.070          & 0.564    & 0.525 & 0.556          & 0.709    \\
        mistral\_zslcot\#0                  & \textbf{0.900} & 0.184          & 0.591    & 0.542 & 0.592          & 0.714    \\
        gemma2\_fsl\#0                      & \textbf{0.765} & 0.326          & 0.564    & 0.545 & 0.574          & 0.655    \\
        qwen2\_zsl\#1                       & \textbf{0.922} & 0.163          & 0.574    & 0.542 & 0.566          & 0.701    \\ \hline
        \end{tabular}}
        \caption{Real dataset}
        \label{tab:subtable2_b}
    \end{subtable}

    \caption{Evaluation of LLMs using their best-performing prompts on the synthetic and seed datasets.}
    \label{tab:model_performance}
\end{table}

A notable strength of almost all models is recall, which measures how effectively the model identifies true positives - in this case, inclusive labels. In addition, Table \ref{tab:model_performance} shows that \textit{gpt-4o-mini} performs well in terms of specificity and precision on real seed data, demonstrating the potential of this latest compact model from OpenAI for language detection tasks.

\subsection{Evaluation of text length and target position}
\begin{figure}[h]
\includegraphics[width=\textwidth]{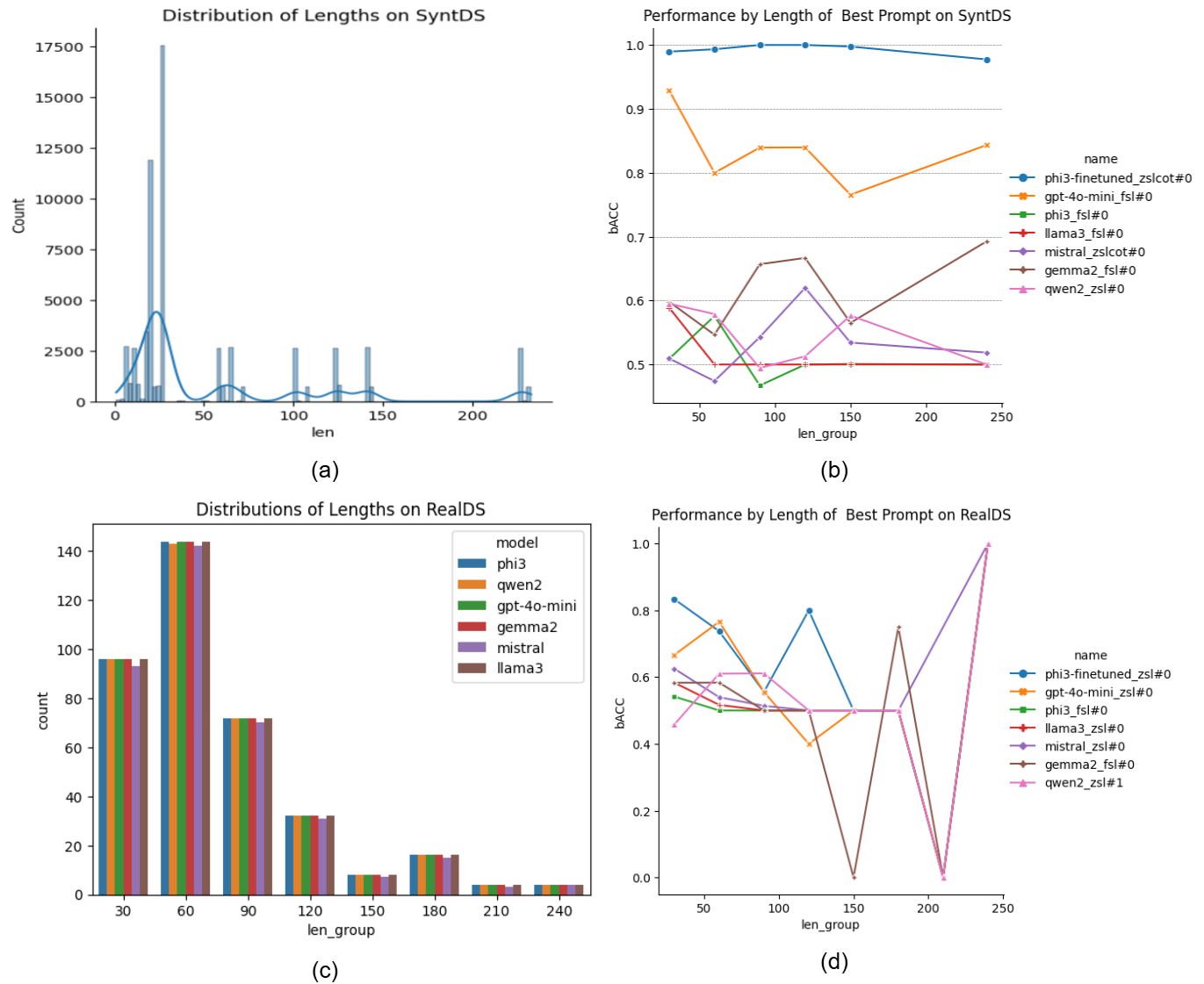}
\caption{The length distribution of the synthetic and seed datasets (a, c) and the performance of the best prompt across different lengths of the synthetic and seed data (b, d)} \label{fig7}
\end{figure}

To explore the relationship between text length and model responses, we categorized the responses into different length groups based on word count. The texts ranged up to 240 words, with the highest concentration in the group of approximately 30-word texts (Figure \ref{fig7}(a)). 
To assess the impact of text length on model performance, we evaluated the accuracy of each model within these groups, generating a line plot that provided valuable insights. Notably, with the synthetic dataset, accuracy remained stable across all models as text length increased, with no significant decrease (Figure \ref{fig7}(b)). 

However, a clear performance gap remained between \textit{phi3-finetuned}, \textit{gpt-4o-mini}, and the other models, highlighting their superior capabilities. 
While the real-world dataset contains a more varied distribution of text lengths, covering both short and medium-length phrases (as shown in Figure \ref{fig7}(c)), the accuracy of each model varies differently depending on the length of the input. This suggests that there is no clear correlation between text length and model performance (Figure \ref{fig7}(d)).

\begin{figure}[ht]
  \centering
    \includegraphics[width=\textwidth]{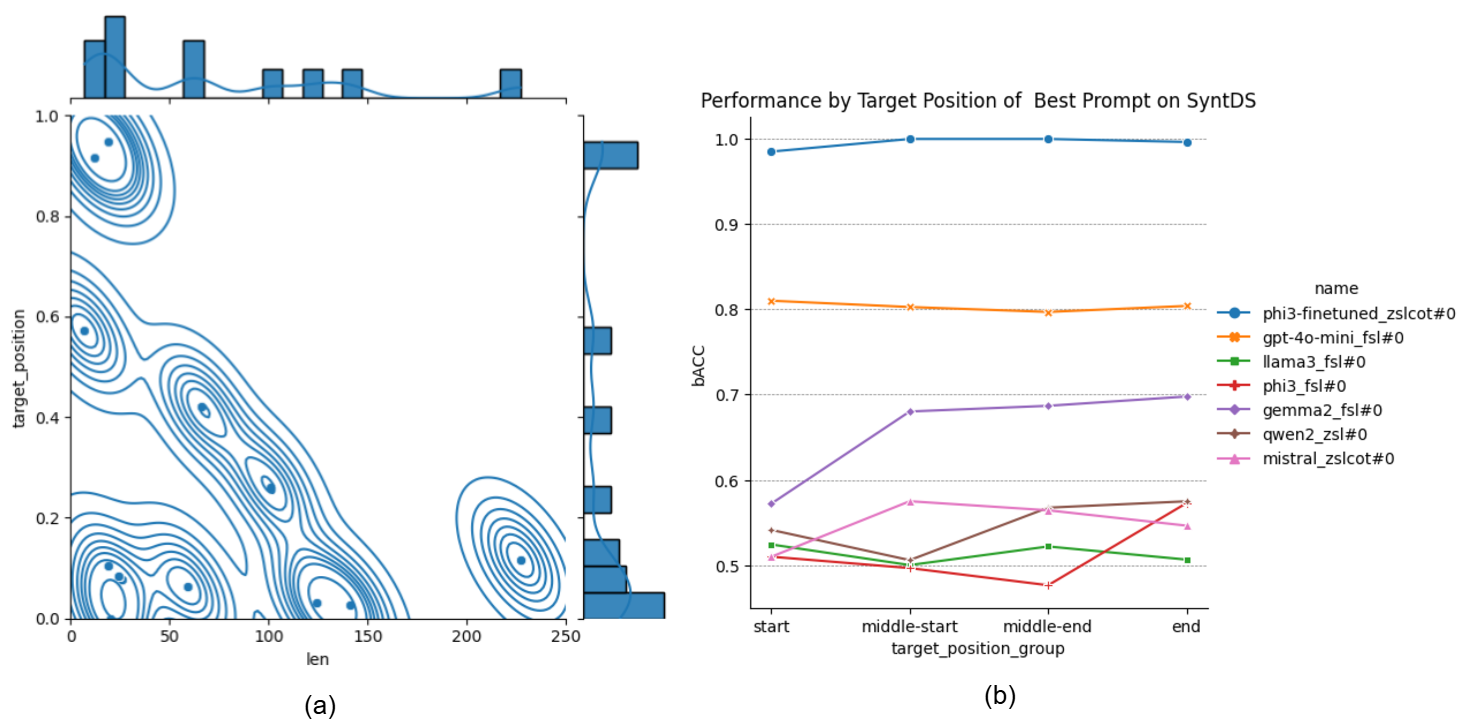}
    \caption{Relationship between target position and text length in synthetic data (a) and the performance of the best prompts based on target position in synthetic data (b).}
    \label{fig9}
\end{figure}

The synthetic dataset also allows for an evaluation of the target words within the phrases, specifically examining their position (start, middle, end) and whether the model's response correctly identifies the target. The plot illustrating the relationship between target position and input length shows that target identification is less reliable in longer texts, particularly when the target appears at the end of the phrase (Figure\ref{fig9}(a)). When analysing the effect of target position on model performance, no significant differences were observed as the target position changed, except for \textit{Gemma2} which showed some variation (Figure\ref{fig9}(b)). 

\section{Conclusion} \label {sec:conclusion}
In this paper, we propose a comprehensive pipeline for generating and evaluating synthetic data using LLMs. Our approach involves extracting sentences from real data and systematically replacing job titles or adjectives with alternatives that have different grammatical endings, adhering to Italian language rules. The resulting synthetic dataset was used to fine-tune \textit{Phi3}, a model from Microsoft. We evaluated the performance of this fine-tuned model against six other pre-trained models. The results demonstrate that the LLM fine-tuned on our synthetic data outperformed the others, achieving superior performance even on real test datasets.

In the future, we plan to extend this methodology for generating synthetic data to domains beyond job descriptions. Additionally, we intend to fine-tune other recent and advanced LLMs, such as \textit{GPT-4o1} and \textit{Gemini}, on this synthetic data to enable comparisons between various fine-tuned models as well.
%
%
%
\newpage
\bibliographystyle{splncs04}

\end{document}